\documentclass[preprint,12pt]{elsarticle}

\usepackage{amsmath,amssymb,amsfonts}
\usepackage{amsthm}
\usepackage{algorithmic}
\usepackage{graphicx}
\usepackage{textcomp}

\usepackage{makecell}
\usepackage{booktabs}
\usepackage{multirow}

\usepackage{array}
\usepackage{hyperref}
\biboptions{sort&compress}
\usepackage{subcaption}

\usepackage{xcolor} 
\newif\ifcolormode
\colormodetrue 

\newcommand{\R}{\mathbb{R}}

\newcommand{\K}{\mathcal{K}}
\newcommand{\x}{\mathbf{x}}
\newcommand{\uvec}{\mathbf{u}}
\newcommand{\F}{\mathcal{F}}

\newcommand{\q}{\mathbf{q}}
\newcommand{\thet}{\boldsymbol{\theta}}
\newcommand{\RNum}[1]{\uppercase\expandafter{\romannumeral #1\relax}}

\newtheorem{theorem}{Theorem}

\newtheorem{definition}{Definition}
\newtheorem{remark}{Remark}

\newcommand\blfootnote[1]{%
  \begingroup
  \renewcommand\thefootnote{}\footnote{#1}%
  \addtocounter{footnote}{-1}%
  \endgroup
}

\journal{Robotics and Autonomous Systems}

\begin{document}

\begin{frontmatter}



\title{\texorpdfstring{\vspace*{-1.9cm}}{}Safe Multi-Robotic Arm Interaction \\via 3D Convex Shapes}


\author{Ali Umut Kaypak\fnref{label1}\corref{cor1}}
\ead{ak10531@nyu.edu}
\author{Shiqing Wei\fnref{label1}}
\ead{shiqing.wei@nyu.edu}
\author{Prashanth Krishnamurthy\fnref{label1}}
\ead{prashanth.krishnamurthy@nyu.edu}
\author{Farshad Khorrami\fnref{label1}} 
\ead{khorrami@nyu.edu}
\cortext[cor1]{Corresponding Author}
\affiliation[label1]{organization={Electrical and Computer Engineering Department, Tandon School of Engineering, New York University},
            city={Brooklyn},
            postcode={11201}, 
            state={NY},
            country={USA}}

\begin{abstract}
Inter-robot collisions pose a significant safety risk when multiple robotic arms operate in close proximity. We present an online collision avoidance methodology leveraging High-Order Control Barrier Functions (HOCBFs) constructed for safe interactions among 3D convex shapes to address this issue. While prior works focused on using Control Barrier Functions (CBFs) for human–robotic arm and single-arm collision avoidance, we explore the problem of collision avoidance between multiple robotic arms operating in a shared space. In our methodology, we utilize the proposed HOCBFs as centralized and decentralized safety filters. These safety filters are compatible with many nominal controllers and ensure safety without significantly restricting the robots' workspace. A key challenge in implementing these filters is the computational overhead caused by the large number of safety constraints and the computation of a Hessian matrix per constraint. We address this challenge by employing numerical differentiation methods to approximate computationally intensive terms. The effectiveness of our method is demonstrated through extensive simulation studies and real-world experiments with Franka Research 3 robotic arms. The project video is available at this \href{https://youtu.be/EePHtPMhON8}{link}. \blfootnote{
  \copyright\ 2025. This manuscript version is made available under the CC-BY-NC-ND 4.0 license 
  \url{https://creativecommons.org/licenses/by-nc-nd/4.0/}. 
  \newline
  The formal publication is available at DOI: \href{https://doi.org/10.1016/j.robot.2025.105263}{https://doi.org/10.1016/j.robot.2025.105263}.
}
\end{abstract}



\end{frontmatter}

\section{Introduction}
The deployment of robots in unstructured dynamic environments necessitates effective collision prevention mechanisms. Various approaches have been proposed for this purpose, including Model Predictive Control (MPC)-based safety strategies \cite{sathya2018embedded,zhang2020optimization} and CBF-based methods \cite{wei2024collision, wang2017safety, kim2024safety, GONCALVES2024104601, cavorsi2023multi, mestres2024distributed}. CBF-based methods have received attention because of their ability to enforce forward invariance of safety sets while maintaining computational efficiency in many scenarios. Leveraging these properties, recent research has extensively explored their application in inter-robot collision avoidance, particularly for safe navigation in mobile robot systems \cite{wang2017safety, kim2024safety, GONCALVES2024104601, cavorsi2023multi}. However, despite the importance of multi-robotic arm interaction in specific applications \cite{smith2012dual,yan2021decentralized}, CBF-based safety certificates for ensuring safe multi-robotic arm interaction remain largely unexplored. This work addresses this gap by investigating the application of CBFs that enhance safety during multi-robotic arm interactions and collaborative tasks.

CBFs are often used as constraints in Quadratic Programming (QP) to enforce safety \cite{ames2019control}.  The QP aims to keep the control input as close to the nominal controller as possible while obeying the safety constraints. To generalize this framework for systems whose relative degree with respect to the constraint function is higher than one, HOCBFs were introduced \cite{tan2021high, xiao2021high}. Our methodology is built on this concept by employing HOCBFs for collision avoidance among 3D convex shapes, as introduced in \cite{wei2024collision}. We extend the use of HOCBFs from single robotic arm collision avoidance to inter-robotic arm collision avoidance by utilizing them in pairwise robotic arm dynamics.

The key challenge in utilizing  HOCBFs constructed for safe interactions among 3D convex shapes in multi-robotic arms, compared to a single arm, is the increased computational overhead. This arises from calculating more constraints at every sampling time. Additionally, in multi-robot settings, all the 3D convex shapes used in the HOCBFs are dynamic. This further intensifies the computational burden. We address this issue by approximating computationally intensive terms via numerical derivative filters. Our contributions in this paper are summarized as follows: 
\begin{itemize}
    \item Developing HOCBFs that ensure robot safety in multi-robotic arm settings and utilizing them in centralized and decentralized paradigms;
    \item Addressing the computational challenges of deploying these HOCBFs by estimating the computationally intensive terms with numerical derivative filters;
    \item Validating the proposed method through extensive simulations and real-world experiments.
\end{itemize}
\section{Related works} 
\textbf{CBFs in inter-robot collision avoidance:} The use of CBFs to prevent collisions in multi-robot systems is an active  research area, with various methodologies proposed for different scenarios \cite{wang2017safety, kim2024safety, GONCALVES2024104601, cavorsi2023multi, mestres2024distributed}. For instance, centralized and decentralized CBF-based safety filters for planar mobile robots were introduced in \cite{wang2017safety}. A CBF-based methodology was developed for the safe control of drone swarms  in \cite{GONCALVES2024104601,GKT24}. In addition, a distributed controller for the safe navigation of mobile robots was proposed in \cite{mestres2024distributed}. Similarly,  HOCBFs were employed to generate safety-assured trajectories for multiple legged robots in  \cite{kim2024safety}. However, extending these approaches to the multi-robotic arm setting is not straightforward. Although some work \cite{shi2024safe, shi2025distributed} applied CBFs to ensure safe human-multiple robot arm interaction, they did not address possible collisions between robot bodies. This is the first work applying CBFs for collision avoidance among multiple robotic arms. 

\textbf{Inter-robotic arm collision avoidance:} Collision handling between robotic arms can be categorized into online \cite{wong2021motion, afaghani2015line, gafur2022dynamic} and offline \cite{chang1994collision, chen2024multiobjective} methods. Offline methods predefine commands and trajectories, while online methods address collisions in real time. Various online strategies have been explored. For example, collision maps were utilized to detect potential collisions and schedule command execution times while preserving the path in \cite{afaghani2015line}. Nonlinear MPC was employed for online trajectory planning, incorporating collision avoidance into the optimization constraints in \cite{gafur2022dynamic}. Additionally, a motion planning method based on Soft Actor-Critic was proposed in \cite{wong2021motion} to prevent collisions in dual-robotic arms. 

In contrast to these approaches, we introduce a CBF-based method designed as a seamless filtering layer compatible with any nominal controller. Unlike these methods, our approach provides reactive collision avoidance. It modifies the nominal controller's inputs (and thus the offline-generated trajectories) only when necessary to prevent potential collisions, thereby preserving the intended high-level task execution. Moreover, our method complements offline collision avoidance approaches, preventing robot collisions that may arise due to unexpected changes in the task environment after the initial trajectory has been planned.

\section{Notations}
The set of real numbers and the set of natural numbers are denoted by \( \mathbb{R} \) and \( \mathbb{N} \), respectively. $\mathbb{R}^n$ and $\mathbb{R}^{m \times n}$ represent real-valued $n$-dimensional column vectors and real-valued $m$-by-$n$ matrices, respectively. The $n$-by-$n$ identity matrix is denoted by \( \mathbf{I}_{n\times n} \). A function is \( \mathcal{C}^k \) if it is \( k \)-times differentiable with a continuous \( k \)th derivative. A continuous function $\Gamma: [0, a) \rightarrow [0, \infty)$, $a > 0$ is a class $\K$ function if it is strictly increasing and $\Gamma(0) = 0$. For a $\mathcal{C}^2$ scalar function $h: \mathbb{R}^n \rightarrow \mathbb{R}$, $\nabla_{\mathbf{x}} h \in \mathbb{R}^{k}$ represents the gradient of $h$ with respect to the vector $\mathbf{x} \in \mathbb{R}^{k}$, $k \leq n$, and $\nabla^2_{\mathbf{x}} h \in \mathbb{R}^{k \times k}$ represents the Hessian of the function with respect to the vector $\mathbf{x}$. For a vector-valued $\mathcal{C}^1$ function $f: \mathbb{R}^{q} \rightarrow \mathbb{R}^m$, $\frac{\partial f}{\partial x}$ represents the partial derivative of $f$, and it is an $m$-by-$k$ matrix, where $\mathbf{x} \in \mathbb{R}^k$ and $k \leq q$. $L_f h(x)$ stands for the Lie derivative of a scalar function $h$ with respect to a vector field $f$ with $m = n$.  The boundary of a set $A$ is represented by $\partial A$. For a column vector \( \x \in \R^n \) and a row vector \( \mathbf{y} \in \R^{1 \times n} \), \( \x_{[k]} \) and \( \mathbf{y}_{[k]} \) denote the \( k \)th element of the vectors \( \x \) and \( \mathbf{y} \), respectively. Similarly, \( \x_{[k:l]} \) and \( \mathbf{y}_{[k:l]} \), where \( k < l \leq n \), represent the subvector \( \begin{bmatrix} \x_{[k]} & \hdots & \x_{[l]} \end{bmatrix}^T \) and the subvector \( \begin{bmatrix} \mathbf{y}_{[k]} & \hdots & \mathbf{y}_{[l]}\end{bmatrix} \), respectively.

\section{Background}
\subsection{High order control barrier functions}
Consider an affine control system in the form
\begin{equation}
    \dot{\mathbf{x}} = f(\mathbf{x}) + g(\mathbf{x})\mathbf{u}, \label{eq:sys}
\end{equation}
where $f: \R^n \rightarrow \R^{n}$ and $g: \R^n \rightarrow \R^{n \times q}$ are Lipschitz $\mathcal{C}^1$ functions with the state vector $\mathbf{x} \in \R^n$ and input vector $\mathbf{u} \in \mathcal{U} \subset \R^q$, where $\mathcal{U}$ denotes a closed set of possible control input values (taking into account physical constraints). For a given $\mathcal{C}^r$ function $h(\mathbf{x}): \R^n \rightarrow \R$ and $r$ class $\K$ functions $\Gamma_k(\cdot)$, $k \in \{1, \ldots, r\}$, a sequence of functions and their superlevel sets are defined as follows:
\begin{equation}
\begin{aligned}
    \psi_0(\mathbf{x}) &= h(\mathbf{x}), \\
    \psi_k(\mathbf{x}) &= \frac{d}{dt}\psi_{k-1}(\mathbf{x})  + \Gamma_k\left(\psi_{k-1}(\mathbf{x}) \right), \quad k \in \{1, \ldots, r\}, \\
    C_k &= \{\mathbf{x} : \psi_{k-1}(\mathbf{x}) \ge 0\}, \quad k \in \{1, \ldots, r\}.
\end{aligned}
\label{eq:psi_definitions}
\end{equation}
\begin{definition}[Time-invariant HOCBFs \cite{xiao2021high}]
\label{def:hocbf}
    Let $C_k, \; k \in \{1, \ldots, r\}$ and $\psi_k(\x), \; k \in \{0, \ldots, r\}$ be defined by (\ref{eq:psi_definitions}). A function $h : \mathbb{R}^n \to \mathbb{R}$ is a time-invariant HOCBF of relative degree $r$ for system (\ref{eq:sys}) if there exist differentiable class $\mathcal{K}$ functions $\Gamma_k, \; k \in \{1, \ldots, r\}$ such that:
\begin{equation}
    \label{eq:cbf}
    \sup_{\mathbf{u} \in \mathcal{U}} \big[L_f \psi_{r-1}(\mathbf{x}) + L_g \psi_{r-1}(\mathbf{x}) \mathbf{u}+ \Gamma_r(\psi_{r-1}(\mathbf{x}))\big] \geq 0
    \end{equation}
for all $\x \in C_1 \cap \cdots \cap C_{r}$.
\end{definition}

Since this paper uses time-invariant HOCBFs, we provide the definition specifically for the time-invariant case rather than the general formulation. Additionally, in Definition~\ref{def:hocbf}, we incorporate the $\psi_{r-1}$ notation within the supremum operator, which is equivalent to
\begin{equation}
    \begin{aligned}
        L_f^r h(\mathbf{x}) + L_g L_f^{r-1} h(\mathbf{x}) \mathbf{u} + \sum_{i=1}^{r-1} L_f^i (\Gamma_{r-i} \circ \psi_{r-i-1})(\mathbf{x}) + \Gamma_r (\psi_{r-1}(\mathbf{x})),
    \end{aligned}
\end{equation}
as the relative degree of $h$ with respect to system (\ref{eq:sys}) is $r$.
\begin{theorem}[Theorem 4 in \cite{xiao2021high} for time-invariant HOCBFs]
\label{thm:HOCBF}
Given an HOCBF $h$ from Definition \ref{def:hocbf}, and with the associated sets $C_k$, $k \in \{1, \hdots, r\}$ defined by (\ref{eq:psi_definitions}), if $x(t_0) \in \mathcal{C}:= C_1 \cap \cdots \cap C_{r}$, then any Lipschitz continuous controller $\uvec(t) \in K_{\text{HOCBF}}(\mathbf{x}) := \{ \mathbf{u} \in \mathcal{U} : L_f \psi_{r-1}(\mathbf{x})
+ L_g \psi_{r-1}(\mathbf{x}) \mathbf{u} + \Gamma(\psi_{r-1}(\mathbf{x})) \geq 0 \} \ \forall t \geq t_0$ renders the set $\mathcal{C}$ forward invariant for the system (\ref{eq:sys}).
\end{theorem}
\subsection{Convex shapes for defining safety in robotic arms}
We utilize 3D convex shapes, specifically ellipsoids, to construct collision bodies encompassing the links and end effectors of robotic arms. We use scaling functions to represent these collision bodies as their 1-sublevel sets in \(\mathbb{R}^3\). 

\begin{definition}[Scaling functions \cite{wei2024diffocclusion}] \label{def:scaling}
A class \(\mathcal{C}^2\) function \(\mathcal{F}_A : \mathbb{R}^{n_p} \times \mathbb{R}^{n_\theta} \to \mathbb{R}\) 
is called a scaling function (with parameters \(\boldsymbol{\theta} \in \mathbb{R}^{n_\theta}\)) for a closed set 
\(A \subset \mathbb{R}^{n_p}\) with non-empty interior if \(\mathcal{F}_A\) is convex in the first argument and
\begin{equation}
A = \left\{\mathbf{p} \in \mathbb{R}^{n_p} \mid \mathcal{F}_A(\mathbf{p}, \boldsymbol{\theta}) \leq 1 \right\}.
\end{equation}
\end{definition}

For a given convex set \(A \subset \mathbb{R}^{n_p}\), the parameter \(\boldsymbol{\theta}\) determines the position and the orientation of the set in \(\mathbb{R}^{n_p}\), while the parameter \(\mathbf{p}\) is used to evaluate the function, satisfying \(\mathcal{F}_A(\mathbf{p}, \boldsymbol{\theta}) \leq 1\) for \(\mathbf{p} \in A\). Consequently,
\begin{equation}
\mathbf{p} \in \partial A(\boldsymbol{\theta}) \quad \Longrightarrow \quad \mathcal{F}_A(\mathbf{p}, \boldsymbol{\theta}) = 1.
\end{equation}

Let \(A\) and \(B\) be convex sets in \(\mathbb{R}^{n_p}\) with \(A \cap B = \emptyset\), and let their corresponding scaling functions be \(\mathcal{F}_A(\mathbf{p},\boldsymbol{\theta})\) and \(\mathcal{F}_B(\mathbf{p},\boldsymbol{\theta})\), respectively. In \cite{wei2024collision}, determining the minimum level set of \(\mathcal{F}_A(\mathbf{p},\boldsymbol{\theta})\) that intersects \(B\) is formulated as the following optimization problem:
\begin{equation}
\label{eq:opt}
\begin{aligned}
    \alpha^\star(\boldsymbol{\theta}) 
    &= \min_{\mathbf{p} \in \mathbb{R}^{n_p}} \mathcal{F}_A(\mathbf{p}, \boldsymbol{\theta}) \\
    &\text{s.t.} \quad \mathcal{F}_B(\mathbf{p}, \boldsymbol{\theta}) \leq 1.
\end{aligned}
\end{equation}
Although there are alternative formulations, we choose the approach in \cite{wei2024collision} because it facilitates the use of torque control in robotic arms. Let $ \mathbf{p}^\star(\boldsymbol{\theta}) $ denote the optimal solution to (\ref{eq:opt}), i.e., the argument that solves (\ref{eq:opt}). The following theorem provides sufficient conditions for the differentiability order of \(\alpha^\star(\boldsymbol{\theta})\):
\begin{theorem}[Theorem 2 in \cite{wei2024collision}]
Let \(\mathcal{F}_A\) and \(\mathcal{F}_B\) be scaling functions corresponding to the sets \(A\) and \(B\), respectively, with \(A \cap B = \emptyset\). Assume that one of \(\mathcal{F}_A\) and \(\mathcal{F}_B\) has a positive definite Hessian in \(\mathbf{p}\). If the scaling functions \(\mathcal{F}_A\) and \(\mathcal{F}_B\) are \(\mathcal{C}^{k+1}\) (with \(k \geq 1\)) in \(\mathbf{p}\) and \(\boldsymbol{\theta}\), then \(\alpha^\star\) is \(\mathcal{C}^k\) in \(\boldsymbol{\theta}\).
\end{theorem}


\section{Safe multi-robotic arm interaction}
\label{sec:overall_cbf}
In Sections \ref{sec:construct} and \ref{sec:control}, we introduce HOCBFs for preventing inter-robot collisions and present their application in centralized and decentralized control paradigms, respectively.

\subsection{Construction of HOCBFs for multi-robotic arm setting}
\label{sec:construct}
Consider an environment with several robotic arms and each robotic arm in the environment is indexed by the set $\mathcal{I} \subset \mathbb{N}$. Dynamics of robotic arm $i \in \mathcal{I}$ are written as \cite{siciliano2008robotics}:
\begin{equation}
\label{eq:single_robot_dynamics}
    M_i(\q_i)\ddot{\q}_i + \underbrace{D_i(\q_i, \dot{\q}_i)\dot{\q}_i + F_i(\dot{\q}_i) + g_i(\q_i)}_{\sigma_i(\q_i, \dot{\q}_i)} = \uvec_i
\end{equation}
where $\q_i \in \mathcal{X}_i \subset \R^{n_i}$ and $\uvec_i \in \mathcal{U}_i \subset \R^{n_i}$ are the joint vector and actuator torque vector for robot $i$, respectively. The sets $\mathcal{X}_i$ and $\mathcal{U}_i$ represent the closed sets of the possible joint configurations and control inputs for robot $i$. $M_i(\q_i) \in \R^{n_i \times n_i}$ is the symmetric positive definite mass matrix, $D_i(\q_i, \dot{\q}_i) \in \R^{n_i \times n_i}$ consists of centrifugal and Coriolis terms, $F_i(\dot{\q}_i) \in \R^{n_i}$ is the friction vector, and  $g_i(\q_i) \in \R^{n_i}$ is the gravity vector for robot $i$. By utilizing (\ref{eq:single_robot_dynamics}), the pairwise robot dynamics for two different robot $i$ and $j$ ($i < j$ to impose a unique ordering convention on pairs) can be expressed in a control affine system form as follows:
\begin{equation}
\begin{aligned}
\frac{d}{dt}
\underbrace{
\begin{bmatrix}
\q_i \\ 
\dot{\q}_i \\ 
\q_j \\ 
\dot{\q}_j
\end{bmatrix}}_{\x_{ij}}
=&
\underbrace{
\begin{bmatrix}
\dot{\q}_i \\ 
-M^{-1}_i(\q_i) \sigma_i(\q_i, \dot{\q_i}) \\ 
\dot{\q}_j \\ 
-M^{-1}_j(\q_j) \sigma_j(\q_j, \dot{\q_j})
\end{bmatrix}
}_{f(\x_{ij})} +
\underbrace{
 \begin{bmatrix}
0 & 0 \\ 
M^{-1}_i(\q_i) & 0 \\
0 & 0 \\  
0 & M^{-1}_j(\q_j)
\end{bmatrix}
}_{g(\x_{ij})}
\underbrace{
\begin{bmatrix}
\uvec_i \\ 
\uvec_j
\end{bmatrix}}_{\uvec_{ij}}.
\end{aligned}
\end{equation}
To construct a scaling function for a robot's link or end effector, we follow the construction recipe described in \cite{wei2024collision}. For a given \({}^{\scriptscriptstyle W}\mathbf{p} \in \mathbb{R}^3  \) in the world frame, the point is first transformed into body frame coordinates as ${}^{\scriptscriptstyle B}\mathbf{p} = {}^{\scriptscriptstyle B}\mathbf{R}_{\scriptscriptstyle W} ({}^{\scriptscriptstyle W}\mathbf{p} - {}^{\scriptscriptstyle W}\mathbf{o}_{\scriptscriptstyle B})$ where \({}^{\scriptscriptstyle B}\mathbf{R}_{\scriptscriptstyle W}\) denotes the orientation of the world frame expressed in the body frame, and \({}^{\scriptscriptstyle W}\mathbf{o}_{\scriptscriptstyle B}\) represents the origin of the body frame in the world frame. 

An analytical solution for \eqref{eq:opt} is available when both scaling functions are ellipsoids \cite{rimon1997obstacle}. In contrast, computing the HOCBF coefficients for other 3D convex shapes requires numerically solving \eqref{eq:opt} at every iteration. This process is computationally costly for real-time robot control. Therefore, we use ellipsoids to represent the robot links and end effector. The scaling functions used in the remainder of this paper are of the following form:
\begin{equation}
\label{eq:ellipsoid}
    \F({}^{\scriptscriptstyle W}\mathbf{p}) = ({}^{\scriptscriptstyle B}\mathbf{R}_{ \scriptscriptstyle W} ({}^{\scriptscriptstyle W}\mathbf{p} - {}^{\scriptscriptstyle W}\mathbf{o}_{\scriptscriptstyle B}) - \boldsymbol{\mu})^T \mathbf{Q}({}^{\scriptscriptstyle B}\mathbf{R}_{\scriptscriptstyle W} ({}^{\scriptscriptstyle W}\mathbf{p} - {}^{\scriptscriptstyle W}\mathbf{o}_{\scriptscriptstyle B}) - \boldsymbol{\mu})
\end{equation}
where $\boldsymbol{\mu} \in \R^3$ represents the ellipsoid center, and $\mathbf{Q} \in \R^{3\times3}$ is a positive definite and symmetric matrix that determines the ellipsoid's principal axes and scaling along each axis.


Consider two distinct robot arms, indexed by $ i $ and $ j $ ($ i < j $), and let \( k \) and \( l \) denote a link or end effector of robots \( i \) and \( j \), respectively. The link centers in the world frame and their quaternions are given by \( \mathbf{o}_{i_k}, \mathbf{o}_{j_l} \in \mathbb{R}^3 \) and \( \boldsymbol{\xi}_{i_k}, \boldsymbol{\xi}_{j_l} \in \mathbb{R}^4 \), respectively. Using these link centers and orientations, an ellipsoid scaling function is defined to represent each link. 

Defining $\thet_{i_kj_{l}}:= \begin{bmatrix}
    \mathbf{o}_{i_k}^T & \boldsymbol{\xi}_{i_k}^T & \mathbf{o}_{j_{l}}^T & \boldsymbol{\xi}_{j_{l}}^T
\end{bmatrix}^T \in \R^{14}$, we obtain the HOCBF and the $\psi$s in (\ref{eq:psi_definitions}) as follows:
\begin{equation}
\label{eq:hocbf_def}
\begin{aligned}
\psi_0^{i_kj_{l}}(\x_{ij}) &= h_{i_kj_l}(\x_{ij}) = \alpha^\star(\thet_{i_kj_{l}}) - \alpha_0, \:\: \alpha_0 > 1 ,\\
\psi_w^{i_kj_{l}}(\x_{ij}) &=  \frac{d}{dt}\psi_{w-1}^{i_kj_{l}}(\x_{ij})  + \Gamma_w\left( \psi_{w-1}^{i_kj_{l}}(\x_{ij})\right) ,  \quad  w \in \{1, 2\}.
\end{aligned}
\end{equation}

By the definition of $\alpha^\star(\thet)$ in~(\ref{eq:opt}), the collision bodies of the pair $(i_k, j_l)$ do not intersect if $\alpha^\star(\thet_{i_kj_{l}})$ is greater than $1$. Therefore, setting the parameter $\alpha_0 > 1$ in~(\ref{eq:hocbf_def}) ensures that the pair $(i_k, j_l)$ of robot links or end-effectors remain collision-free within the safe set defined by the proposed HOCBF. Specifically, the parameter $\alpha_0$ defines the safety margin between these bodies by shifting the boundary of the feasible region. Larger values of $\alpha_0$ correspond to more conservative safety distances, while smaller values allow robots to operate closer to each other.

By applying the chain rule, the time derivative of \( h_{i_kj_l}(\x_{ij}) \) is given by  
\begin{equation}
\label{eq:h_derivative}
    \dot{h}_{i_kj_l}(\x_{ij}) = \left(\nabla_{\thet} \alpha^{\star} (\thet_{i_kj_l})\right)^T \dot{\thet}_{i_kj_l}
\end{equation}
where  
\begin{equation}
    \dot{\thet}_{i_kj_l} = \mathbf{T}_{ij}^{kl} \, \boldsymbol{\Omega}_{ij}^{kl} \, \dot{\q}_{ij}
\end{equation}
with  
\begin{equation}
\begin{aligned}
    \mathbf{T}_{ij}^{kl} = \begin{bmatrix}
        \mathbf{I}_{3 \times 3} & 0 & 0 & 0 \\
        0 & 0.5 \Pi(\boldsymbol{\xi}_{i_k}) & 0 &0 \\
        0 & 0 & \mathbf{I}_{3 \times 3} & 0 \\
        0 & 0 & 0 & 0.5 \Pi(\boldsymbol{\xi}_{j_l})
    \end{bmatrix}, \\
    \boldsymbol{\Omega}_{ij}^{kl} = \begin{bmatrix}
    J_{i_k}(\q_i) & 0 \\ 
    0 & J_{j_l}(\q_j)
    \end{bmatrix}, \quad
    \dot{\q}_{ij} = \begin{bmatrix}
        \dot{\q}_i \\ \dot{\q}_j
    \end{bmatrix}.
\end{aligned}
\end{equation}
with the matrix \( \Pi(\boldsymbol{\xi}) \) defined as  
\begin{equation}
\Pi(\boldsymbol{\xi}) =
\begin{bmatrix}
    \boldsymbol{\xi}_w & \boldsymbol{\xi}_z & -\boldsymbol{\xi}_y \\
    -\boldsymbol{\xi}_z & \boldsymbol{\xi}_w & \boldsymbol{\xi}_x  \\
    \boldsymbol{\xi}_y & -\boldsymbol{\xi}_x & \boldsymbol{\xi}_w \\
    -\boldsymbol{\xi}_x & -\boldsymbol{\xi}_y & -\boldsymbol{\xi}_z 
\end{bmatrix},
\end{equation}
and $J_{i_k}(\q_i) \in \R^{6\times n_i}$ and  $J_{j_{l}}(\q_j) \in \R^{6\times n_j}$ are the geometric Jacobian matrices for the link $k$ in robot $i$ and $l$ in robot $j$. 

Utilizing (\ref{eq:hocbf_def}) and (\ref{eq:h_derivative}), along with the chain rule, we obtain $L_g \psi_{1}^{i_kj_{l}}(\x_{ij})$ and $L_f \psi_1^{i_kj_{l}}(\x_{ij})$ as follows:

\begin{equation}
    L_g \psi_{1}^{i_kj_{l}}(\x_{ij}) = \left(\nabla_{\boldsymbol{\theta}}\alpha^{\star}(\thet_{i_kj_{l}})\right)^T\mathbf{T}_{ij}^{kl} \boldsymbol{\Omega}_{ij}^{kl}\boldsymbol{\Upsilon}_{ij},
\end{equation}
\begin{equation}
\label{eq:appr_foundation}
    \begin{aligned}
        L_f \psi_1^{i_kj_l}(\x_{ij}) =   \dot{\thet}_{i_kj_{l}}^T \left(\nabla^2_{\thet}\alpha^{\star} (\thet_{i_kj_{l}})\right) \dot{\thet}_{i_kj_{l}} +\dot{\Gamma}_1(\psi_0^{i_kj_l}(\x_{ij})) 
        \\ + \left(\nabla_{\boldsymbol{\theta}}\alpha^{\star} (\thet_{i_kj_{l}})\right)^T\left(  \dot{\mathbf{T}}_{ij}^{kl}\boldsymbol{\Omega}_{ij}^{kl} \dot{\q}_{ij} +\mathbf{T}_{ij}^{kl} \dot{\boldsymbol{\Omega}}_{ij}^{kl}\dot{\q}_{ij} - \mathbf{T}_{ij}^{kl} \boldsymbol{\Omega}_{ij}^{kl}\boldsymbol{\Upsilon}_{ij}\boldsymbol{\sigma}_{ij}\right),
    \end{aligned}
\end{equation}
where 
\begin{equation}
    \boldsymbol{\Upsilon}_{ij} =       \begin{bmatrix}
 M^{-1}_i(\q_i) & 0 \\
 0 & M^{-1}_j(\q_j)
 \end{bmatrix},\, \boldsymbol{\sigma}_{ij} = \begin{bmatrix}
     \sigma_i(\q_i, \dot{\q_i}) \\\sigma_j(\q_j, \dot{\q_j})
 \end{bmatrix}. 
\end{equation}

$L_g \psi_{1}^{i_kj_{l}}(\x_{ij})$ and $L_f \psi_1^{i_kj_{l}}(\x_{ij})$ can be computed analytically using sensor data, robot kinematics, and robot dynamics.

\subsection{Centralized-decentralized HOCBF filters}
\label{sec:control}
The centralized HOCBF filter employs a single shared filter to process all robots' nominal input signals collectively, whereas the decentralized filter applies individual safety filters to each robot in the environment. The notion of multi-robot safety in this paper is formally defined in Definition~\ref{def:safety}. As our focus is on inter-robotic arm collision avoidance, this definition does not account for self-collisions. However, analogous HOCBFs for self-collision avoidance can also be formulated by using single-robot dynamics in place of the pairwise dynamics in the multi-robot HOCBF construction described in Sec.~\ref{sec:construct}.  Similarly, the proposed framework can be extended to dynamic obstacles in the environment by treating each obstacle as an additional agent. Pairwise HOCBFs can then be constructed between the obstacle components and the robot links and end-effectors, and the resulting constraints can be integrated into the safety filters.

\begin{definition}
\label{def:safety}
    Consider an environment with multiple robotic arms indexed by the set $\mathcal{I} \subseteq \mathbb{N}$, with their dynamics given by (\ref{eq:single_robot_dynamics}). For each robot $i \in \mathcal{I}$, let $\mathcal{P}_i \subseteq \mathbb{N}$ index the links and end-effector of robot $i$. For all $i, j \in \mathcal{I}$ with $i < j$, and for all $k \in \mathcal{P}_i$ and $l \in \mathcal{P}_j$, define the sets:
    \begin{equation}
        C^{i_kj_l}_w := \{\mathbf{x}_{ij} : \psi_{w-1}^{i_kj_l}(\mathbf{x}_{ij}) \geq 0\}, \quad  w \in \{1, 2\}
    \end{equation}
    where $\psi_w^{i_kj_l}(\mathbf{x}_{ij})$ is defined as in (\ref{eq:hocbf_def}). The multi-robot system is \textbf{safe} if, for all $i, j \in \mathcal{I}$ with $i < j$, and for all $k \in \mathcal{P}_i$ and $l \in \mathcal{P}_j$, we have $\mathbf{x}_{ij}(t) \in C^{i_kj_l}_1 \; \forall t \geq t_0$ whenever $\mathbf{x}_{ij}(t_0) \in C^{i_kj_l}_1$.
\end{definition}

 In Definition~\ref{def:safety}, each set $C^{i_k j_l}_1$ is defined as the superlevel set of $\psi^{i_k j_l}_0(\mathbf{x}_{ij}) = \alpha^\star(\boldsymbol{\theta}_{i_k j_l}) - \alpha_0$, where nonnegative values indicate that the corresponding robot link pair $(i_k, j_l)$ remains safely separated. Definition~\ref{def:safety} enforces this condition for all link and end-effector pairs belonging to different robotic arms. Therefore, in terms of collision avoidance, it defines multi-robotic arm safety. Since the relative degree of our inter-robot safety HOCBFs with respect to the pairwise robot dynamics is two, the higher-order sets $C^{i_k j_l}_2$ are necessary to ensure the forward invariance of the sets $C^{i_k j_l}_1$. Specifically, the sets $C^{i_k j_l}_2$ impose velocity-related conditions, guaranteeing that the safety constraints defined by $C^{i_k j_l}_1$ remain dynamically enforceable under appropriate control inputs.

In the centralized safety filter, a single QP determines each robot's control input, i.e. $\uvec_i$, ensuring compliance with the safety constraints in Definition \ref{def:safety}.  Alongside the multi-robot safety HOCBFs, constraints on joint angles, velocities, and torques for each robot are incorporated into the QP to ensure that the resulting controllers respect the robots’ physical limits.

Given upper and lower joint angle limits $\underline{\q}_i, \overline{\q}_i \in \R^{n_i}$ for robot $i \in \mathcal{I}$, we define the HOCBFs restricting joint angles $\forall m \in \{1, \dots, n_i\}$ as $\psi_{0}^{\underline{\q_i}m}(\x_i) = \q_{i\left[m\right]} - \underline{\q_i}{}_{\left[m\right]}$ and $\psi_{0}^{\overline{\q_i}m}(\x_i) =  \overline{\q_i}{}_{\left[m\right]} - \q_{i\left[m\right]} $. Since joint angle constraints are second-order with respect to the control input, we construct $\psi_{w}^{\underline{\q_i}m}$ and $\psi_{w}^{\overline{\q_i}m}$ for $w \in {1, 2}$ iteratively, following the definition in (\ref{eq:psi_definitions}). The resulting HOCBF constraints for joint angle limits are then given by:
\begin{equation}
    \label{eq:lower_angle}
    L_g \psi_1^{\underline{\q}_im}(\mathbf{x}_i) \uvec_i \geq - L_f  \psi_1^{\underline{\q}_im}(\mathbf{x}_{i}) - \Gamma_2^{\underline{\q}_i}( \psi_1^{\underline{\q}_im}(\mathbf{x}_{i})),
\end{equation}
\begin{equation}
    \label{eq:upper_angle}
    L_g \psi_1^{\overline{\q}_im}(\mathbf{x}_i) \uvec_i \geq - L_f  \psi_1^{\overline{\q}_im}(\mathbf{x}_{i}) - \Gamma_2^{\overline{\q}_i}( \psi_1^{\overline{\q}_im}(\mathbf{x}_{i})).
\end{equation}

Similarly, $\forall m \in \{1,\dots,n_i\}$ we define $\psi_{0}^{\underline{\dot{\q}_i}m}(\x_i)= \dot{\q}_{i\left[m\right]} - \underline{\dot{\q}_i}{}_{\left[m\right]}$ and $\psi_{0}^{\overline{\dot{\q}_i}m}(\x_i) =  \overline{\dot{\q}_i}{}_{\left[m\right]} - \dot{\q}_{i\left[m\right]} $ for the lower and upper joint velocity limits $\underline{\dot{\q}_i}, \overline{\dot{\q}_i} \in \R^{n_i}$. Since joint velocity constraints are first-order with respect to the control input, we construct $\psi_{1}^{\underline{\dot{\q}_i}m}$ and $\psi_{1}^{\overline{\dot{\q}_i}m}$ according to the definition in (\ref{eq:psi_definitions}). The resulting HOCBF constraints for joint angular velocity limits are given by:
\begin{equation}
\label{eq:vel_lower}
    L_g \psi_0^{\underline{\dot{\q}_i}m}(\mathbf{x}_i) \uvec_i \geq - L_f  \psi_0^{\underline{\dot{\q}_i}m}(\mathbf{x}_{i}) - \Gamma_1^{\underline{\dot{\q}_i}}( \psi_0^{\underline{\dot{\q}_i}m}(\mathbf{x}_{i})),
\end{equation}
\begin{equation}
\label{eq:vel_upper}
    L_g \psi_0^{\overline{\dot{\q}}_im}(\mathbf{x}_i) \uvec_i \geq - L_f  \psi_0^{\overline{\dot{\q}}_im}(\mathbf{x}_{i}) - \Gamma_1^{\overline{\dot{\q}}_i}( \psi_0^{\overline{\dot{\q}}_im}(\mathbf{x}_{i})).
\end{equation}

To formulate the centralized filter, let $\mathbf{u}_{\mathrm{nom}_i}$ denote the nominal control for robot $i \in \mathcal{I}$. Let $\underline{\uvec_i}$ and $\overline{\uvec_i}$ represent lower and upper torque limits for robot i. Define the stacked nominal control input and the stacked control input as $ \uvec_{\mathrm{nom}} := \begin{bmatrix}
        \uvec_{\mathrm{nom}_i}^T \mid i \in \mathcal{I}
    \end{bmatrix}^T$, $ \uvec := \begin{bmatrix}
        \uvec_i^T \mid i \in \mathcal{I}
    \end{bmatrix}^T$, respectively. Given a positive definite matrix \(\mathbf{P}\), \emph{centralized-HOCBF-QP} is formulated as:
\begin{equation}
\label{eq:centralized}
    \begin{aligned}
        &\min_{\uvec} \quad (\uvec - \uvec_{\mathrm{nom}})^T\mathbf{P}(\uvec - \uvec_{\mathrm{nom}}) \\
    & \text{s.t.} \: \forall i, j \in \mathcal{I}, \text{with}\: i < j,  \forall k \in \mathcal{P}_i \: \text{and}\: \forall l \in \mathcal{P}_j: \\
        &  L_g  \psi_1^{i_kj_l}(\mathbf{x}_{ij}) \mathbf{u}_{ij}  \geq - L_f  \psi_1^{i_kj_l}(\mathbf{x}_{ij}) - \Gamma_2( \psi_1^{i_kj_l}(\mathbf{x}_{ij})).\\
        & \forall i \in \mathcal{I} \quad \forall m \in \{1, \dots, n_i\}: \:  \overline{\uvec_i}{}_{[m]} \geq \uvec_{i[m]} \geq  \underline{\uvec_i}{}_{[m]},\\
        &  \text{Angle constraints for robot $i$, joint $m$ in (\ref{eq:lower_angle})--(\ref{eq:upper_angle})}, \\
        & \text{Velocity constraints for robot $i$, joint $m$ in (\ref{eq:vel_lower})--(\ref{eq:vel_upper})}.
    \end{aligned}
\end{equation}
The solution to this optimization problem is unstacked in the same order as $\uvec_{\mathrm{nom}}$ and provided to the respective robots in the environment. By Theorem~\ref{thm:HOCBF}, if the multi-robot system starts in a configuration where
\begin{equation}
\label{eq:start}
\x_{ij}(t_0) \in \bigcap_{\substack{w = 1, 2\\ k \in \mathcal{P}_i,l \in \mathcal{P}_j 
}}C^{i_kj_l}_w, \quad \forall i, j \in \mathcal{I}, \, i < j  
\end{equation}
and (\ref{eq:centralized}) is feasible at all times, then the proposed filter ensures safety for all \( t \ge t_0 \). 

The matrix $\mathbf{P}$ in the \emph{centralized-HOCBF-QP} serves as a task-priority weighting matrix, where larger diagonal entries correspond to robots with higher task priorities, and smaller entries to those with lower priorities.

The computational cost of solving the QP problem increases with the number of optimization parameters, which grows as the number of robots increases. Therefore, \emph{centralized-HOCBF-QP} faces scalability challenges due to the computational burden of solving a large-scale optimization problem for real-time control. \emph{Decentralized-HOCBF-QP} provides a viable solution for this problem by enabling each robot to independently filter its own nominal control input in parallel during task execution, at the cost of suboptimal solutions. Given a positive definite matrix $\textbf{P}_i$, \emph{decentralized-HOCBF-QP} for robot $i \in \mathcal{I}$ is formulated as: 
\begin{equation}
\label{eq:decentralized}
    \begin{aligned}
    & \min_{\substack{\uvec_i} \in \R^{n_{i}}} \quad ( \uvec_i - \uvec_{\mathrm{nom}_i})^T\textbf{P}_i ( \uvec_i - \uvec_{\mathrm{nom}_i}) \\
    & \text{s.t.} \: \forall j \in \mathcal{I}, \text{with}\: i < j,  \forall k \in \mathcal{P}_i \: \text{and}\: \forall l \in \mathcal{P}_j: \\
    & L_g  \psi_1^{i_kj_l}(\mathbf{x}_{ij})_{\left[1:n_i\right]} \mathbf{u}_{i}  \geq - c_{ij} (L_f  \psi_1^{i_kj_l}(\mathbf{x}_{ij}) + \Gamma_2( \psi_1^{i_kj_l}(\mathbf{x}_{ij}))).\\
    & \: \forall j \in \mathcal{I}, \text{with}\: i > j,  \forall k \in \mathcal{P}_i \: \text{and}\: \forall l \in \mathcal{P}_j: \\
    & L_g  \psi_1^{j_li_k}(\mathbf{x}_{ji})_{\left[n_j+1:n_j +n_i\right]} \mathbf{u}_{i}  \geq - c_{ij} (L_f  \psi_1^{j_li_k}(\mathbf{x}_{ji}) + \Gamma_2( \psi_1^{j_li_k}(\mathbf{x}_{ji}))).\\
    & \: \forall m \in \{1, \dots, n_i\}: \:  \overline{\uvec_i}{}_{[m]} \geq \uvec_{i[m]} \geq  \underline{\uvec_i}{}_{[m]},\\
    & \:  \text{Angle constraints for robot $i$, joint $m$ in (\ref{eq:lower_angle})--(\ref{eq:upper_angle})}, \\
        & \:  \text{Velocity constraints for robot $i$, joint $m$ in (\ref{eq:vel_lower})--(\ref{eq:vel_upper})}.
    \end{aligned}
\end{equation}

In decentralized filtering, priority among different robot tasks is not introduced through the cost function, as it includes only a single robot's control. Instead, it is established via the scalar $c_{ij}$, which appears in the QP  constraints. The scalar $c_{ij} \in [0, 1]$ represents the responsibility of robot $i$ in the HOCBF constraint that is shared with robot $j$, and it satisfies the condition $c_{ij} +c_{ji}= 1$. Assuming each robot has access to other robots' states, the coefficients of the inequality constraints can either be computed independently by each robot or computed collectively and shared. The choice between these approaches depends on the specific multi-robot system setting.

The relationship between the centralized and decentralized filters can be explained as follows. Assuming that $\mathbf{P}$ in \emph{centralized-HOCBF-QP} has a block-diagonal structure, where each block $\mathbf{P}_i$ corresponds to robot $i$, the cost function in \emph{centralized-HOCBF-QP} can be written as $\sum_{i \in \mathcal{I}} (\uvec_i - \uvec_{\mathrm{nom}_i})^T \mathbf{P}_i (\uvec_i - \uvec_{\mathrm{nom}_i})$. To separate the optimization constraints for each robot, the multi-robot safety HOCBF constraints in \emph{centralized-HOCBF-QP} are decoupled through the introduction of responsibility coefficients $c_{ij}$. This decomposition, together with the independence of the cost function terms across robots, isolates each robot’s control input in the optimization problem from those of the other robots. As a result, each robot can solve its own QP locally and in parallel, referred to as \emph{decentralized-HOCBF-QP}, enforcing its share of the safety constraints while considering only its control input.

As with \emph{centralized-HOCBF-QP}, Theorem~\ref{thm:HOCBF} guarantees that if the multi-robot system starts in a configuration where condition~(\ref{eq:start}) holds and (\ref{eq:decentralized}) is feasible for all \( i \in \mathcal{I} \) at all times, then \emph{decentralized-HOCBF-QP} ensures safety for all \( t \ge t_0 \). However, the solution to \emph{decentralized-HOCBF-QP} is suboptimal compared to the \emph{centralized-HOCBF-QP} solution due to the decoupling of the shared multi-robot safety constraints through the responsibility coefficients \( c_{ij} \) in (\ref{eq:decentralized}). This also reduces the feasibility set of the control inputs, which may render (\ref{eq:decentralized}) infeasible for some \( i \in \mathcal{I} \). To mitigate this issue, a relaxation method was introduced in decentralized QPs in prior work \cite{wang2017safety} for first-order CBFs. Following this approach, we propose \emph{decentralized-HOCBF-relaxed-QP}  and establish its theoretical foundation for HOCBFs in Theorem~\ref{thm:relaxedHOCBF}.
\begin{theorem}
\label{thm:relaxedHOCBF}
Let \( h \) be a time-invariant HOCBF, and consider \( \psi_{k-1} \) with the corresponding safety sets \(C_k \) for \( k \in \{1, \ldots, r\} \) as defined in (\ref{eq:psi_definitions}). Suppose that the class $\K$ functions $\Gamma_k$s used to define $\psi_{k}$s are locally Lipschitz. Furthermore, let \( \phi(t) \) be continuous and satisfy \( \phi(t) \geq 1 \) for all \( t \geq t_0 \). Then, any Lipschitz continuous controller $\mathbf{u} : \mathbb{R}^n \to \mathbb{R}^q$ such that $\mathbf{u}(\mathbf{x}) \in K_{\text{RHOCBF}}(\mathbf{x}) := \{ \mathbf{u} \in \mathcal{U} : L_f \psi_{r-1}(\mathbf{x})
+ L_g \psi_{r-1}(\mathbf{x}) \mathbf{u} + \phi(t)\Gamma_r(\psi_{r-1}(\mathbf{x})) \geq 0 \}$ will render the set $\mathcal{C} := \bigcap_{k=1}^r C_k$ forward invariant for the system (\ref{eq:sys}).
\end{theorem}
\begin{proof}
The proof is adapted along the lines of Theorem  \RNum{5}.1 in \cite{wang2017safety}. First, consider the system $\dot{z} = -\Gamma_r(z)$ where $\Gamma_r$ is a class $\K$ function. The solution of the system is $z(t)=\sigma(z(0), t)$, where $\sigma(k,s)$ is a class $\K \mathcal{L}$ function and $\frac{\partial}{\partial s} \sigma(k,s) = -\Gamma_r(\sigma(k,s))$ \cite{khalil1996nonlinear}.  Then the solution of the relaxed system, i.e. $\dot{\hat{z}} = -\phi(t) \Gamma_r(\hat{z})$, is found as \begin{equation}\label{eq:relaxed_sys}
    \hat{z}(t) = \sigma\left(\hat{z}(0), \int_0^t \phi(\tau)d\tau\right). 
    \end{equation} The solution can be verified by 
\begin{equation}
\begin{aligned}
\frac{d\left(\sigma\left(\hat{z}(0), \int_0^t \phi(\tau)d\tau\right) \right)}{dt}
= \frac{\partial\sigma(\hat{z}(0),s) }{\partial s}  
\frac{d\left( \int_0^t \phi(\tau)d\tau \right)}{dt} \\
= - \Gamma_r\left(\sigma\left(\hat{z}(0), \int_0^t \phi(\tau)d\tau\right)\right)\phi(t)
= - \Gamma_r\left(\hat{z}\right)\phi(t) = \dot{\hat{z}}
\end{aligned}
\end{equation}
and 
\begin{equation}
    \sigma\left(\hat{z}(0), \int_0^0 \phi(\tau)d\tau\right) = \sigma\left(\hat{z}(0), 0\right) = \hat{z}(0).
\end{equation}
The solution is unique because $\Gamma_r$ is locally Lipschitz and $\phi$ is continuous. Since $\mathbf{u}(\mathbf{x}) \in K_{\text{RHOCBF}}(\mathbf{x})$, we have $\dot{\psi}_{r-1}(\x) = L_f \psi_{r-1}(\mathbf{x})
+ L_g \psi_{r-1}(\mathbf{x}) \mathbf{u} \geq -\phi(t)\Gamma_r(\psi_{r-1}(\x))$. From (\ref{eq:relaxed_sys}) and Comparison Lemma \cite{khalil1996nonlinear}, it follows that
\begin{equation}
\psi_{r-1}(\mathbf{x}) \geq  \sigma\left(\psi_{r-1}(\x(0)), \int_0^t \phi(\tau)d\tau\right).
\end{equation} 
Assuming $\x(0) \in \mathcal{C}$, the definition of $\mathcal{C}$ implies $\psi_{k}(\x(0)) \geq 0,\, 0 \leq k \leq r-1 $. Since $\phi(t) \geq 1$, $\int_0^t \phi(\tau)d\tau$ is monotonically increasing with $t$ and $\int_0^t \phi(\tau)d\tau \rightarrow \infty$ as $t \to \infty$. By the properties of $\K \mathcal{L}$ functions, for a fixed $u \geq 0$, $\sigma(u, v) \rightarrow 0 $ as $v \rightarrow \infty$  and as it is decreasing, $\psi_{r-1}(\mathbf{x}) \geq 0$ for all $t\geq 0$. Given the recursive definition $ \psi_k(\x)=  \dot{\psi}_{k-1}(\x) +  \Gamma_k \left(\psi_{k-1}(\x) \right)$, if $\psi_k(\x) \geq 0 $, then $\dot{\psi}_{k-1}(\x) \geq -\Gamma_k({\psi}_{k-1}(\x))$. Since $\Gamma_k$ is a class $\mathcal{K}$ function and $\mathbf{x}(0) \in \mathcal{C}$, it follows by induction that $\psi_k(\mathbf{x}) \geq 0$ for $0 \leq k \leq r-1$. This establishes that $K_{\text{RHOCBF}}(\mathbf{x})$ renders the set forward-invariant for system (\ref{eq:sys}).
\end{proof}

As the only constraint on $\phi(t)$ is that it must be continuous, we follow \cite{wang2017safety} and introduce it as an optimization parameter in the decentralized QPs. The optimization problem includes the hyperparameter $\lambda$, which controls the trade-off between increasing $\phi(t)$ and adhering to the original objective, and is formulated for robot $i \in \mathcal{I}$ as follows:
\begin{equation}
\label{eq:relaxed}
    \begin{aligned}
        &\underset{\uvec_i \in \R^{n_i},\, \phi_{i_kj_l} \geq 1}{\min} \quad  
        ( \uvec_i - \uvec_{\mathrm{nom}_i})^T\textbf{P}_i ( \uvec_i - \uvec_{\mathrm{nom}_i}) +
        \lambda \sum_{\substack{j \in \mathcal{I},\, j\neq i \\ k \in \mathcal{P}_i,\, l \in \mathcal{P}_j} } (\phi_{i_kj_l}-1)^2  \\
    & \text{s.t.} \: \forall j \in \mathcal{I}, \text{ with } i < j,  \forall k \in \mathcal{P}_i \text{ and } \forall l \in \mathcal{P}_j: \\
    & L_g  \psi_1^{i_kj_l}(\mathbf{x}_{ij})_{\left[1:n_i\right]} \mathbf{u}_{i}  \geq - c_{ij} L_f  \psi_1^{i_kj_l}(\mathbf{x}_{ij}) - c_{ij}\phi_{i_kj_l}\Gamma_2( \psi_1^{i_kj_l}(\mathbf{x}_{ij})).\\
    & \: \forall j \in \mathcal{I}, \text{with}\: i > j,  \forall k \in \mathcal{P}_i \: \text{and}\: \forall l \in \mathcal{P}_j: \\
    & L_g  \psi_1^{j_li_k}(\mathbf{x}_{ji})_{\left[n_j+1:n_j +n_i\right]} \mathbf{u}_{i}  \geq - c_{ij} L_f  \psi_1^{j_li_k}(\mathbf{x}_{ji}) - c_{ij} \phi_{i_kj_l} \Gamma_2( \psi_1^{j_li_k}(\mathbf{x}_{ji})).\\
     &  \forall m \in \{1, \dots, n_i\}: \:  \overline{\uvec_i}{}_{[m]} \geq \uvec_{i[m]} \geq  \underline{\uvec_i}{}_{[m]},\\
        &    \text{Angle constraints for robot $i$, joint $m$ in (\ref{eq:lower_angle})--(\ref{eq:upper_angle})}, \\
        &  \text{Velocity constraints for robot $i$, joint $m$ in (\ref{eq:vel_lower})--(\ref{eq:vel_upper})}.
    \end{aligned}
\end{equation}
Theorem~\ref{thm:relaxedHOCBF} guarantees that if the multi-robot system begins in a configuration satisfying condition~(\ref{eq:start}) and (\ref{eq:relaxed}) remains feasible for all \( i \in \mathcal{I} \) at all times, then the \emph{decentralized-HOCBF-relaxed-QP} maintains safety for all \( t \geq t_0 \). 

Feedback linearization, which uses joint accelerations as virtual control inputs, is a widely used technique for robotic arm control. Within this framework, the nominal control inputs are defined in terms of accelerations, making it more convenient to formulate safety filters in the acceleration space. In this formulation, we define the QP cost functions in terms of joint accelerations and their nominal values, and replace the torque inputs $\uvec_i$ in the QP constraints with their equivalent dynamic representation $M_i(\q_i)\ddot{\q}_i + \sigma_i(\q_i, \dot{\q}_i)$. Specifically, given a positive definite matrix $\hat{\textbf{P}}_i$, the cost function of \emph{decentralized-HOCBF-QP} and the control input component of the cost function of \emph{decentralized-HOCBF-relaxed-QP} are written as
\begin{equation}
\label{eq:acc_decentralized}
    (\ddot{\q}_i - \ddot{\q}_{\mathrm{nom}_i})^T\hat{\textbf{P}}_i ( \ddot{\q}_i - \ddot{\q}_{\mathrm{nom}_i}),
\end{equation}
where $\ddot{\q}_{\mathrm{nom}_i}$ is the nominal acceleration input. Similarly, given a positive definite matrix $\hat{\textbf{P}}$, the cost function of \emph{centralized-HOCBF-QP} in this formulation is written as 
\begin{equation}
\label{eq:acc_centralized}
    (\ddot{\q} - \ddot{\q}_{\mathrm{nom}})^T\hat{\textbf{P}} ( \ddot{\q} - \ddot{\q}_{\mathrm{nom}}),
\end{equation}
where 
$\ddot{\q}_{\mathrm{nom}} := \begin{bmatrix}
        \ddot{\q}_{\mathrm{nom}_i}^T \mid i \in \mathcal{I}
    \end{bmatrix}^T$ and 
$\ddot{\q} := \begin{bmatrix}
        \ddot{\q}_i^T \mid i \in \mathcal{I}
\end{bmatrix}^T$.

\begin{remark}[QP feasibility in the safety filters]
    In the centralized filter, QP infeasibility, i.e. an empty feasible set, can theoretically arise, although we did not encounter this situation in any experiment. A typical place where it may become infeasible is in highly cluttered environments where a robot's movement is prevented by multiple surrounding robots. Even when the centralized QP remains feasible, one or more QPs in the decentralized filters may become infeasible under the same scenario. This occurs because the decentralized formulation shrinks the feasible sets to decouple the shared problem into independent subproblems. When QP infeasibility is observed,  a fail-safe mode that halts all robots by applying accelerations opposite to their joint velocities is activated. Since the safety filters already bound joint speeds, this braking action stops the robots before contact.  However, activating the fail-safe mode interrupts the high-level task being executed by the nominal controller and requires new high-level planning before resuming operation. In practice, we observed no QP infeasibility with the centralized filter and only a few cases with the decentralized filters; see Sec.~\ref{sec:experiments} for details.
\end{remark}

\section{Computational challenges in computing HOCBFs for multi-robotic arm safety}
\label{sec:approximation}
\begin{figure*}[!ht]
    \centering
\includegraphics[width=\columnwidth]{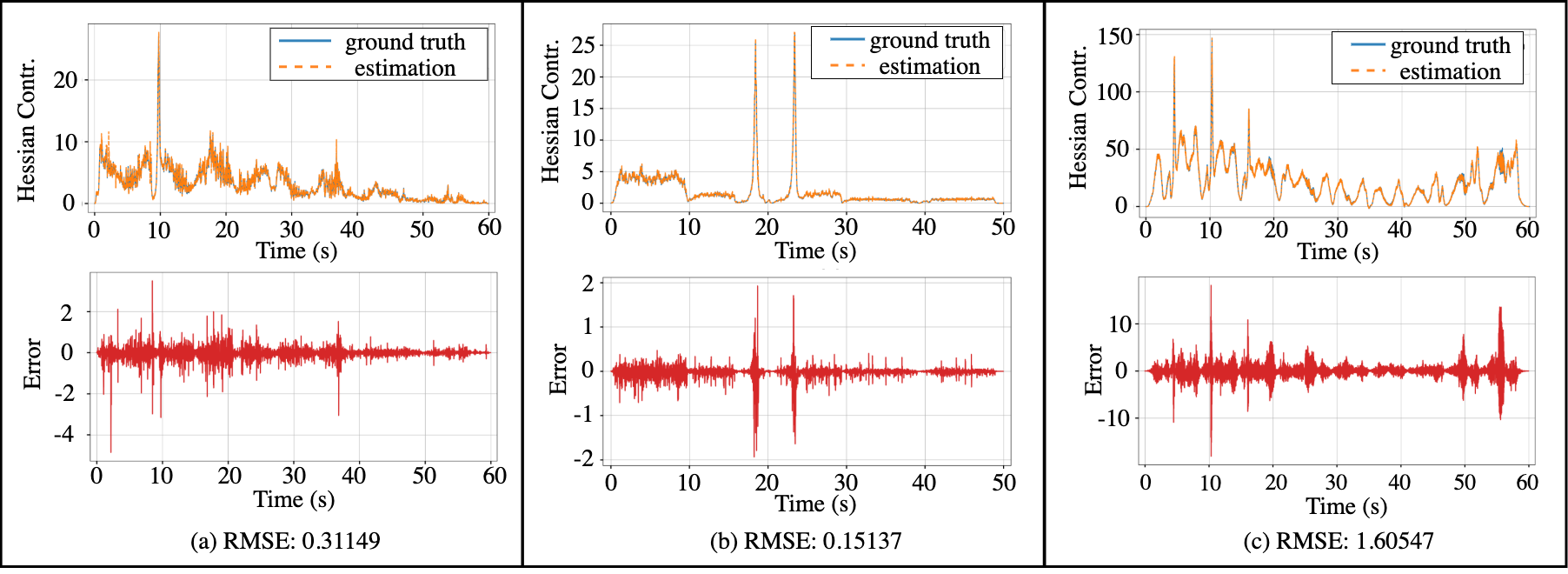}   
    \caption{Three examples of \emph{Hessian contribution} estimation utilizing Savitzky-Golay filter in a real multi-robot environment. The top plots compare the estimated signals to the ground truth, and the bottom plots present the corresponding estimation errors. The root-mean-square error (RMSE) for each case is reported below. The data and ground truth signals were collected from real experiments that use analytically computed \emph{Hessian contributions}.  The estimation was performed offline, simulating an online setting. The window length and polynomial order in the filter are set to 5 and 2, respectively.}
    \label{fig:main_figure}
\end{figure*}

In this section, we address the computational challenges of employing the HOCBFs in multi-robot environments. During the experiments presented in Sec.~\ref{sec:experiments}, we observed that the primary computational overhead in HOCBF computations arises from computing \( L_f \psi_1^{i_kj_l}(\x_{ij}) \). While this overhead does not hinder computation of HOCBFs that prevent a single robot from colliding with static or dynamic obstacles as in \cite{wei2024collision}, it poses significant computational challenges in multi-robot environments due to the large number of HOCBFs required to ensure multi-robot safety. For instance, while 14 HOCBFs are sufficient to protect the 7 links of a robotic arm from two obstacles, 48 HOCBFs are required to protect the last 4 links of the three robotic arms from colliding with each other. Furthermore, the need to define \( \thet \) as a 14-dimensional vector for two moving objects—compared to 7 dimensions for a moving robot and a static obstacle—further amplifies the computational burden of HOCBF computations.

$L_f \psi_1^{i_kj_l}(\x_{ij})$ is computed as in (\ref{eq:appr_foundation}). The primary bottleneck in this computation stems from calculating the Hessian \(\nabla^2_{\thet}\alpha^{\star} (\thet_{i_kj_{l}}) \in \R^{14 \times 14}\). This Hessian is required to compute the term \(\dot{\thet}_{i_kj_{l}}^T\left(\nabla^2_{\thet}\alpha^{\star} (\thet_{i_kj_{l}})\right) \dot{\thet}_{i_kj_{l}}\). We refer to this term as \emph{Hessian contribution} throughout the remainder of the paper. To alleviate the computational burden, we propose approximating  \(\left(\nabla^2_{\thet}\alpha^{\star} (\thet_{i_kj_{l}})\right) \dot{\thet}_{i_kj_{l}} = \frac{d}{dt}\left(\nabla_{\thet}\alpha^{\star} (\thet_{i_kj_{l}})\right)\) via numerical differentiation. This numerical derivative is obtained by using the gradient computed at the current time step together with gradients from previous time steps. The approximated \emph{Hessian contribution} is then obtained by taking the dot product of this numerical derivative with  \(\dot{\thet}_{i_kj_{l}}\). Note that the gradient \(\nabla_{\thet} \alpha^{\star}(\thet_{i_kj_{l}})\) is already computed as part of other terms in the HOCBF formulation, so no additional gradient evaluation is necessary. Since real robot sensor readings can be noisy and susceptible to environmental perturbations, we use a robust numerical differentiation filter, the Savitzky-Golay filter \cite{savitzky1964smoothing}, in our methodology, where its parameters can be tuned according to the noise level.

The Savitzky–Golay filter is characterized by two hyperparameters: the window length and the polynomial order. It fits a polynomial of the chosen order to the most recent window of data using least squares and computes the derivative from the fitted curve. These hyperparameters affect the accuracy of the estimated \emph{Hessian contribution}. An excessively large window length produces overly smoothed and lagged derivatives that fail to capture rapid changes. Conversely, a small window makes the filter sensitive to the noise in the signal. Higher-order polynomials capture complex trends but are more noise-sensitive, while lower orders offer better robustness against noise. In our implementation, we use a second-degree polynomial and select the window length based on the expected noise level. Our proposed safety method operates robustly for a reasonable range of these hyperparameter values, indicating low sensitivity to their precise selection. Nonetheless, inappropriate hyperparameter choices can amplify derivative estimation errors, distort the estimated \emph{Hessian contribution}, and ultimately compromise safety.

Fig.~\ref{fig:main_figure} demonstrates three examples of the \emph{Hessian contribution} estimated using the proposed approximation method alongside their ground truth in an environment with two Franka Research 3 robotic arms. As shown in this figure and supported by the RMSE between the analytically computed and estimated \emph{Hessian contribution}, we conclude that the proposed method achieves sufficient accuracy for estimating the \emph{Hessian contribution} in a real robot environment. Additional evidence for the method’s accuracy is presented in Sec.~\ref{sec:experiments}, where the numerical estimation of the \emph{Hessian contribution} does not compromise the safety of the robots.

\begin{table}[!h]
    \centering
     \caption{Average Time (In Milliseconds) Required to Compute HOCBFs Using the Analytically Computed Hessian Contribution and the Savitzky-Golay Filter Estimation.}
    \label{tab:time_difference}
    \footnotesize
    \begin{tabular}{c c c}
    \toprule
    \# of HOCBFs & Analytical Method & \makecell{Savitzky-Golay Filter\\Estimation} \\
    \midrule
    1 & 4.40 & 1.24 \\
    8 & 4.46 & 1.27 \\
    32 & 10.43 & 2.96 \\
    64 & 17.31 & 4.99 \\
    128 & 30.38 & 8.90 \\
    256 & 56.23 & 16.48 \\
    \bottomrule\addlinespace[1ex]
    \end{tabular}
\end{table}

A comparison of the time required to compute HOCBFs using the analytical and numerical methods is provided in Tab.~\ref{tab:time_difference}. The computations were performed on a computing system equipped with a 24-core Intel i7-13700KF processor and 32 GB of RAM. Parallel computing was utilized, with each HOCBF computation assigned to a separate thread for simultaneous execution. Due to the parallel computing, a linear increase in computation time with the number of HOCBFs is not observed in Tab.~\ref{tab:time_difference}. While the HOCBF computations, excluding the Savitzky-Golay filtering, were implemented in C++ using the \texttt{xtensor} library \footnote{\url{https://xtensor.readthedocs.io}}, the simulations were conducted in Python, employing the \texttt{scipy} \footnote{\url{https://scipy.org}} implementation of the Savitzky-Golay filter. The average in the table was computed over 3600 runs.

As shown in Tab.~\ref{tab:time_difference}, the proposed \emph{Hessian contribution} estimation method significantly reduces computational time. This estimation approach enables scalability of HOCBFs described in Sec.~\ref{sec:overall_cbf} to scenarios involving more than two robotic arms and facilitates higher precision control for two-robot arm setups at sufficiently high update rates.   

\section{Experiments}
\label{sec:experiments}
We designed both simulation and real-world experiments to evaluate the proposed method. In real-world experiments, we demonstrated the applicability of the methodology under real-world conditions and showed the effect of utilizing numerical derivation in \emph{Hessian contribution} on the control update rate, while in simulations, we showcased its scalability to scenarios involving more than two robots and various path-following tasks. Additionally, we exhaustively compared centralized, decentralized, and relaxed-decentralized HOCBF filters in the simulations. In all experiments, feedback linearization was used, and the safety filters were formulated in terms of joint accelerations. An identity matrix was employed for the matrices $\hat{\textbf{P}}_i$ and $\hat{\textbf{P}}$ in (\ref{eq:acc_decentralized}) and (\ref{eq:acc_centralized}). The parameter $c_{ij}$ in the decentralized and relaxed-decentralized filters was set to $0.5$. $\lambda$ in the relaxed-decentralized filter was chosen as $10$. We utilized OSQP \cite{osqp} to solve the QP problems. For the nominal controller, we incorporated an inverse dynamics controller as the primary objective for trajectory-following in the operational space.

In the experiments, we utilized linear class $\K$ functions for multi-robot safety HOCBFs, i.e $\Gamma_w (\psi_w^{i_kj_{l}}(\x_{ij})) = \Gamma_{w} \psi_w^{i_kj_{l}}(\x_{ij})$ in (\ref{eq:hocbf_def}). We set $\Gamma_{1} = 15$ and $\Gamma_{2} = 15$ for these functions. Similarly, we used linear class $\K$ functions for the joint angle and velocity HOCBFs, setting $\Gamma_1^{\underline{\q}_i} = \Gamma_2^{\underline{\q}_i} = \Gamma_1^{\overline{\q}_i} = \Gamma_2^{\overline{\q}_i} = 20$ and $\Gamma_1^{\underline{\dot{\q}_i}} = \Gamma_1^{\overline{\dot{\q}}_i} = 10$. We selected $\alpha_0 = 1.03$ as the parameter in~(\ref{eq:hocbf_def}). This value provided a small but sufficient safety margin to prevent collisions while avoiding excessive conservatism in the experiments.

\subsection{Simulation studies}
\begin{figure}[!h]
    \centering
    \includegraphics[width=0.75\columnwidth]{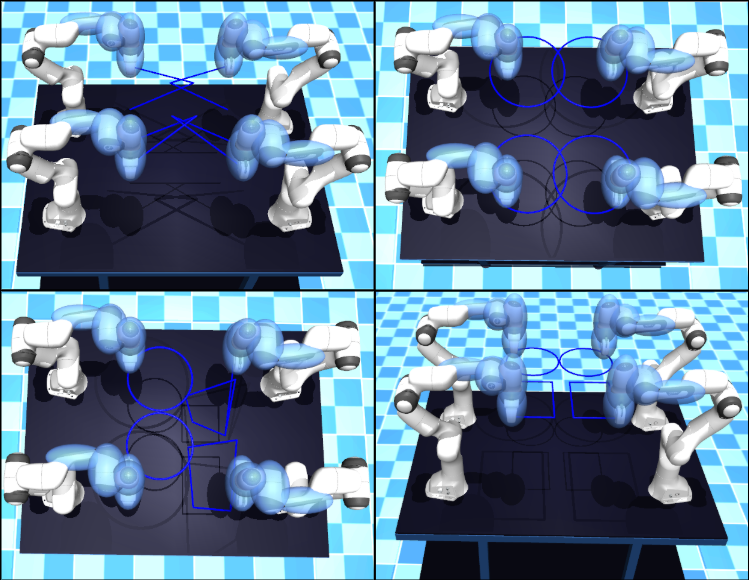}   \caption{Trajectory following examples in the MuJoCo environment. Ellipses covering the robot parts represent the 3D shapes used to define HOCBFs. Dark blue line segments in front of the end-effectors illustrate examples of intersecting trajectories.}
    \label{fig:simulation}
\end{figure}

We conducted our simulation experiments in MuJoCo \cite{todorov2012mujoco}. The simulation environment, consisting of four Franka Research 3 robotic arms with intersecting workspaces, is illustrated in Fig.~\ref{fig:simulation}. Each robot's end effector and last three links were enclosed by four ellipsoids to define HOCBFs, resulting in a total of 96 HOCBFs for multi-robot safety. In the simulation, we assumed that HOCBF calculations were performed in real-time, ensuring no control delays.

\begin{table}[!h]
\centering
\caption{Success rates of the HOCBF filtering methods with computed (Comp.) and estimated (Est.) Hessian Contribution in the 4-robot MuJoCo Environment.}
\label{tab:method_comp}
\footnotesize
\begin{tabular}{@{}c|cccccc@{}}
\toprule
\textbf{Methods}                                                      & \multicolumn{2}{c}{Centralized} & \multicolumn{2}{c}{Decentralized} & \multicolumn{2}{c}{\begin{tabular}[c]{@{}c@{}}Relaxed\\ Decentralized \end{tabular}}  \\
 \cmidrule(lr){2-3} \cmidrule(lr){4-5} \cmidrule(lr){6-7}
\textbf{\begin{tabular}[c]{@{}c@{}}Hessian\\ Contr. \end{tabular}} & Comp.        & Est.        &   Comp.        & Est.  & Comp.        & Est.                \\
\cmidrule(lr){2-7}
\textbf{\begin{tabular}[c]{@{}c@{}}Success\\ Rate\end{tabular}}     & 1 & 1 & 0.96 & 0.96 & 0.98  &   0.98                                       \\ 
\bottomrule
\end{tabular}%
\end{table}

To simulate sensor noise, we added Gaussian noise to the joint angles and velocities: $\mathcal{N}(0, 1\times 10^{-3})$ for joint angles (rad) and $\mathcal{N}(0, 2.5\times 10^{-3})$ for joint angular velocities (rad/s).  This noise also affected derived entities such as ellipsoid frame positions, orientations, and Jacobians. The Savitzky-Golay filter used a window length of 25 and polynomial order of 2. We conducted 100 experiments for each filter with diverse trajectories, including circular and linear paths, where at least two robots would collide without safety measures. Four examples of these trajectories are shown in Fig.~\ref{fig:simulation}. An experiment was considered unsuccessful if the robots collided, or one of the HOCBF-QP problems became infeasible. The success rates for the centralized, decentralized, and relaxed-decentralized filters, using both analytically computed and estimated \emph{Hessian contributions}, are reported in Tab.~\ref{tab:method_comp}. 

\begin{figure*}[!h]
    \centering
\includegraphics[width=0.95\columnwidth]{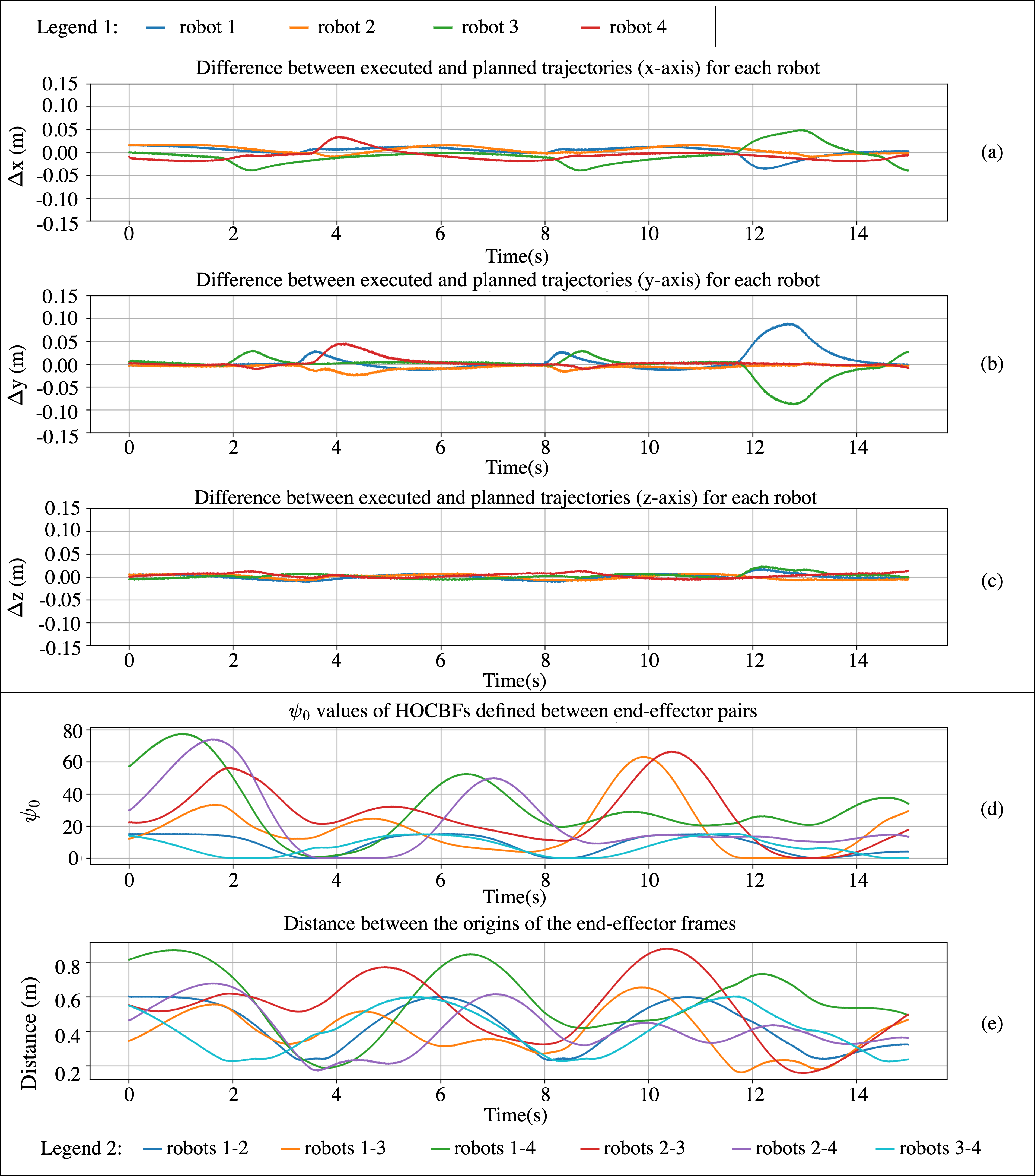}
    \caption{Subplots (a)–(c) show the deviations $\Delta x$, $\Delta y$, and $\Delta z$ between executed and planned trajectories for robots 1–4 over the course of a simulation experiment using centralized filter. Subplot (d) shows the time evolution of the HOCBF values $\psi_{0}$ defined between each pair of end-effectors.  Subplot (e) shows the distance between the origins of the end-effector frames for each robot pair. In subplots (a)–(c), the legend identifies individual robots; in subplots (d)–(e), it identifies the six inter-robot pairs.}
    \label{fig:sim_trajectories}
\end{figure*}

\begin{figure}[!h]
    \centering
    \includegraphics[width=0.98\linewidth]{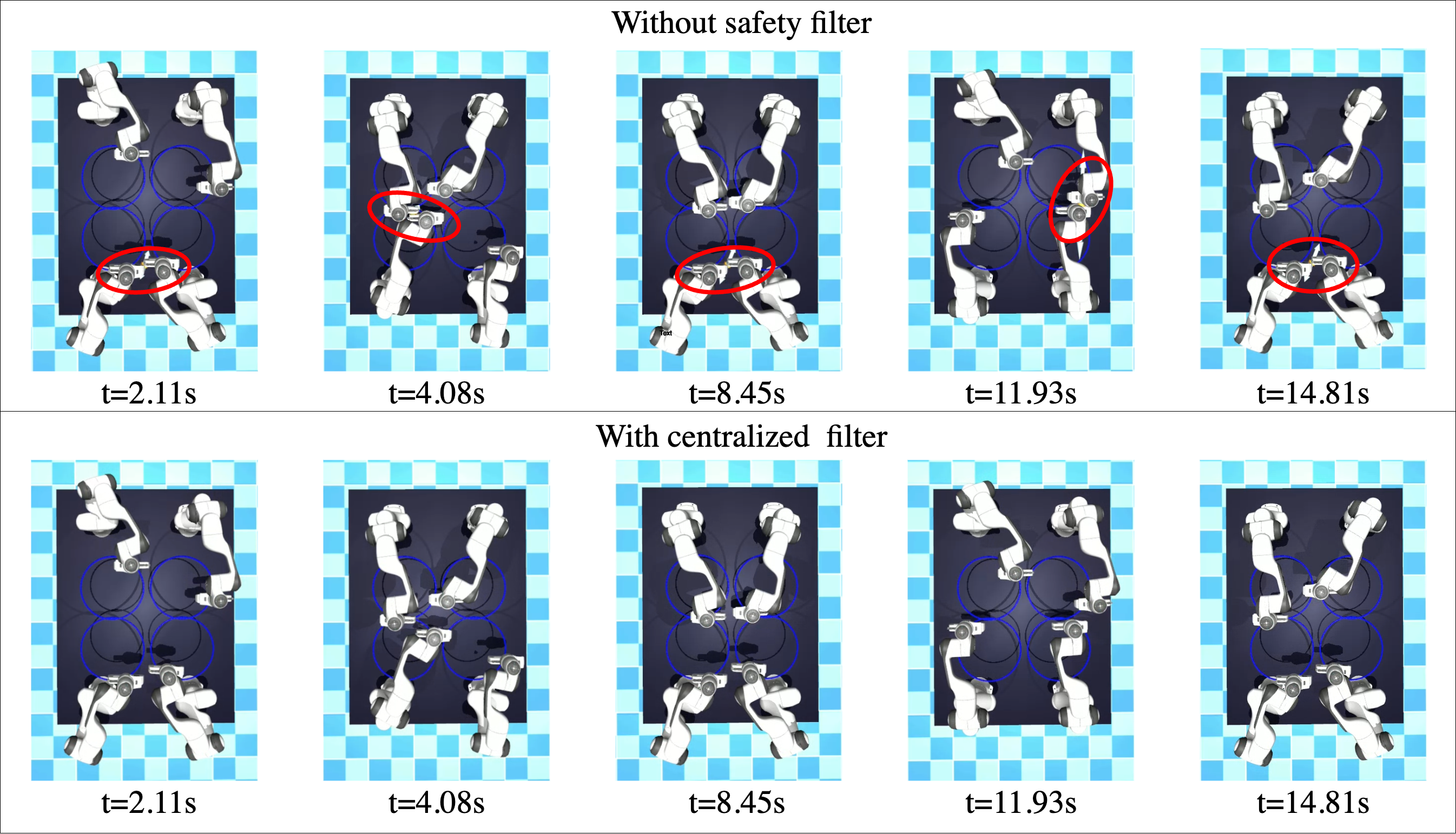}
    \caption{Comparison of the same experiment under the nominal controller (top row), where collisions occur, and with the centralized HOCBF filter (bottom row), which prevents collisions. Red ellipsoids indicate robot collisions.}
    \label{fig:comparative}
\end{figure}

As shown in Tab.~\ref{tab:method_comp}, the estimated \emph{Hessian contribution} preserved the HOCBF safety guarantees, indicating its reliability when analytical computation is infeasible. While the centralized filters succeeded in all experiments, the decentralized filters failed in some cases due to QP infeasibility. Although relaxation in decentralized filtering mitigated this issue in some experiments, it did not fully eliminate it. In all instances where a QP became infeasible, the fail-safe mode successfully halted all robots and prevented collisions. When centralized filters were employed as a backup strategy in these cases, we observed that infeasibility did not arise, allowing the task to continue without interruption. Moreover, although the reduced feasible set in the decentralized filter led to slightly more conservative behavior compared to the centralized filters, task execution remained feasible. The relaxed-decentralized filter helped to alleviate this conservatism.

Fig.~\ref{fig:sim_trajectories} illustrates the differences between the planned and executed trajectories, the $\psi_0$ values of the HOCBFs defined for the end-effector pairs, and the distances between the origins of their frames during a simulation experiment with the centralized filter. The results show that when the $\psi_0$ values are significantly above zero, indicating a safe distance between the robots, the deviations from the planned trajectories remain minimal. In contrast, as $\psi_0$ approaches zero, the executed trajectories diverge from the planned paths to maintain safety, demonstrating the effectiveness of the safety filter in preventing collisions. A comparative visualization of this experiment, highlighting collisions under the nominal controller and the corresponding collision-free outcomes with the centralized filter at the same time instances, is shown in Fig.~\ref{fig:comparative}.

\subsection{Real-world experiments}
\begin{figure}[!h]
    \centering
    \includegraphics[width=0.5\linewidth]{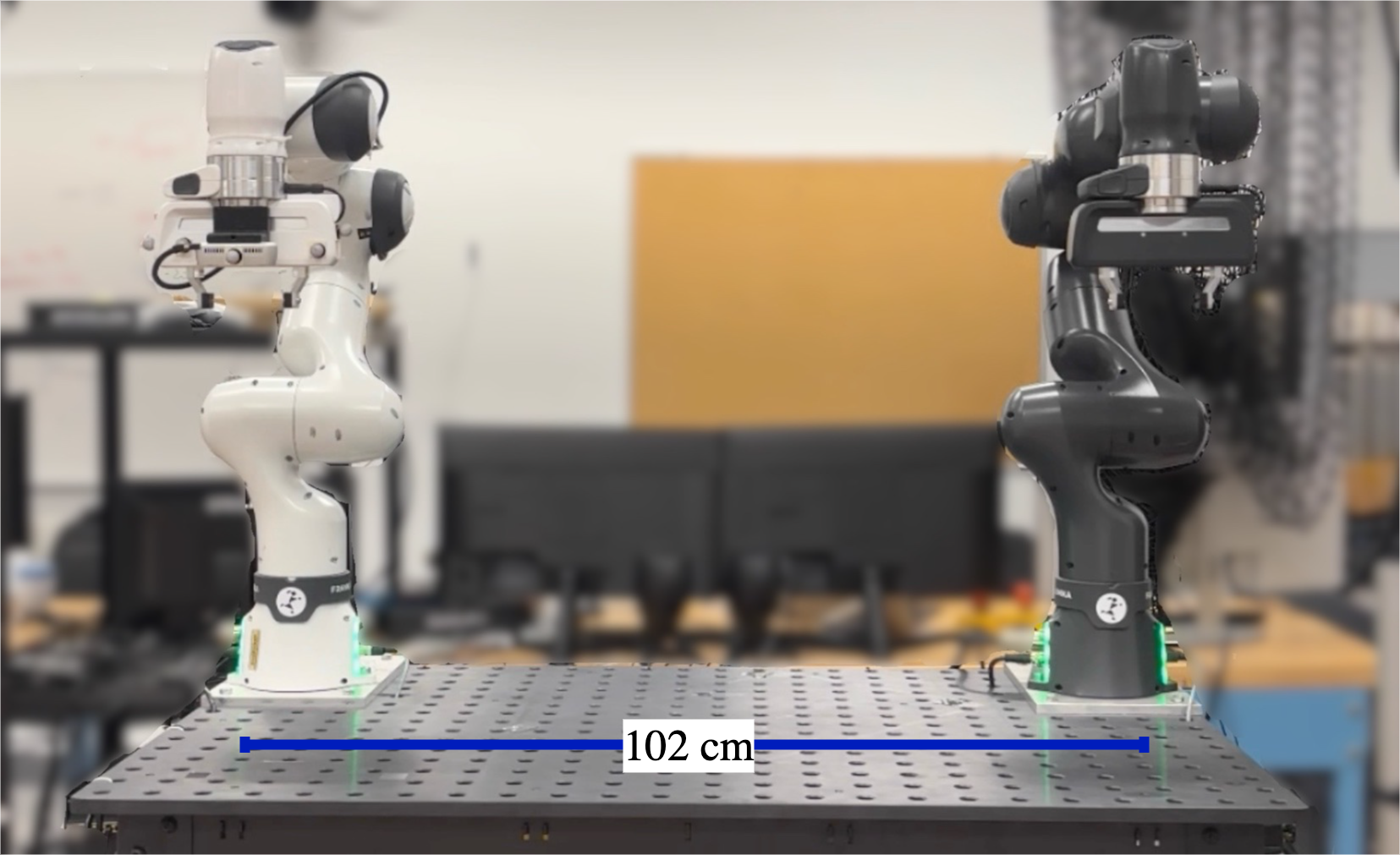}
    \caption{Real-world experimental environment.}
    \label{fig:real_world_experiment environment}
\end{figure}
\begin{figure*}[!h]
    \centering
\includegraphics[width=\columnwidth]{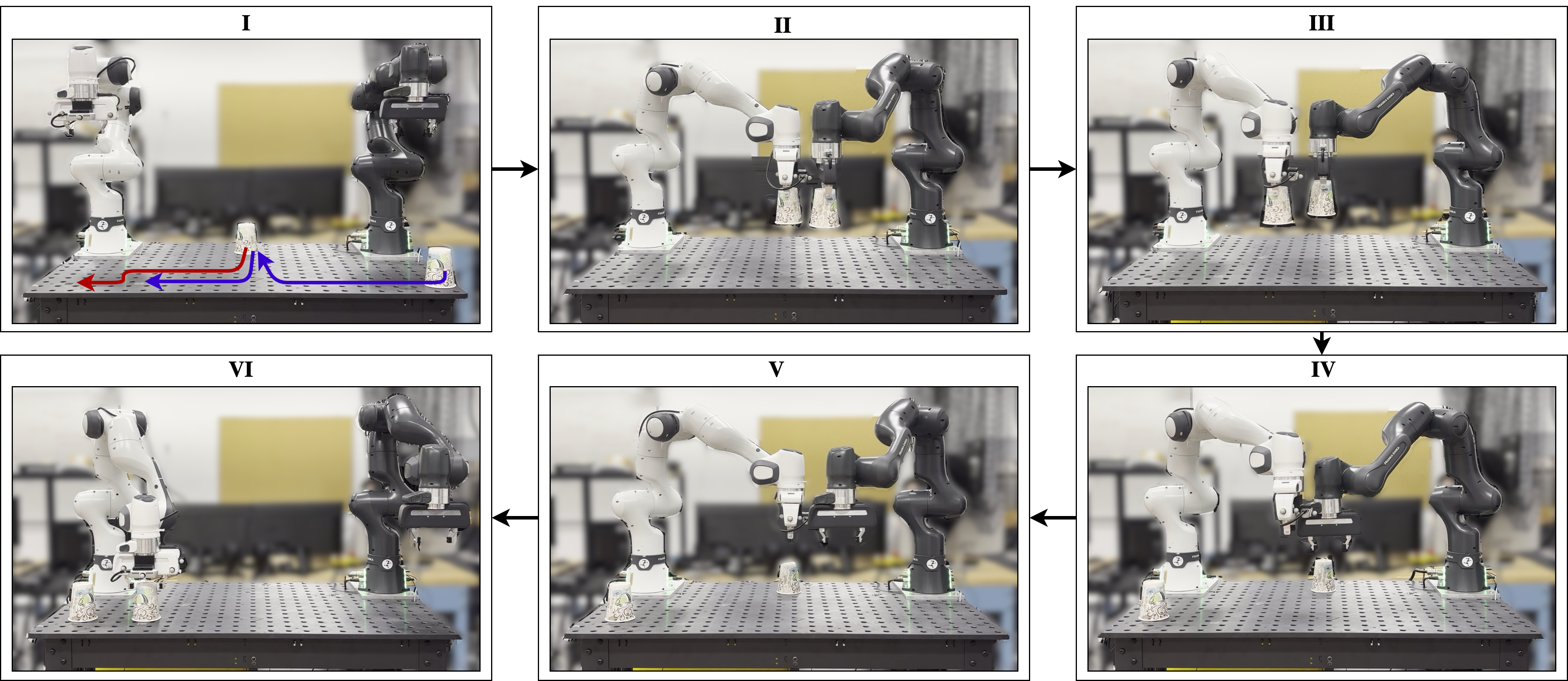}
    \caption{Two-robot pick-and-place example: The white arm moves an object to its final position while the gray arm places another object in the first object's initial spot, which the white arm later relocates. Their trajectories intersect twice (Images \RNum{2} and \RNum{4}), but collisions are prevented by our method.} 
    \label{fig:real_robot}
\end{figure*}
\begin{figure}[!h]
    \centering
    \includegraphics[width=0.7\columnwidth]{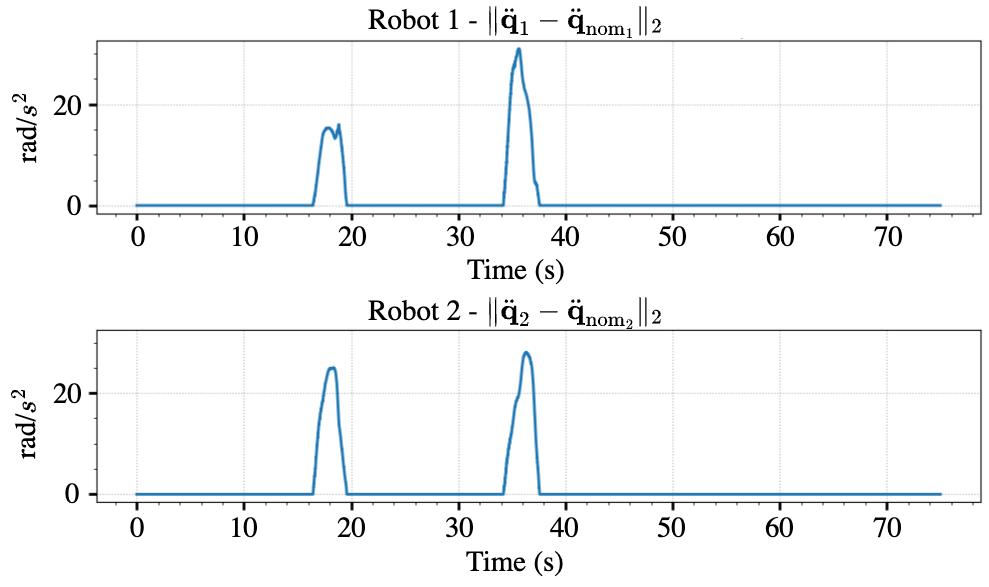}
    \caption{2-norm deviation between the nominal joint acceleration and the safe-filtered joint acceleration vectors for each robot during the pick-and-place task. Peaks correspond to instants when the safety filter intervenes to enforce collision avoidance.}
    \label{fig:nominalvsfiltered}
\end{figure}
\begin{figure}[!h]
    \centering
    \includegraphics[width=0.7\columnwidth]{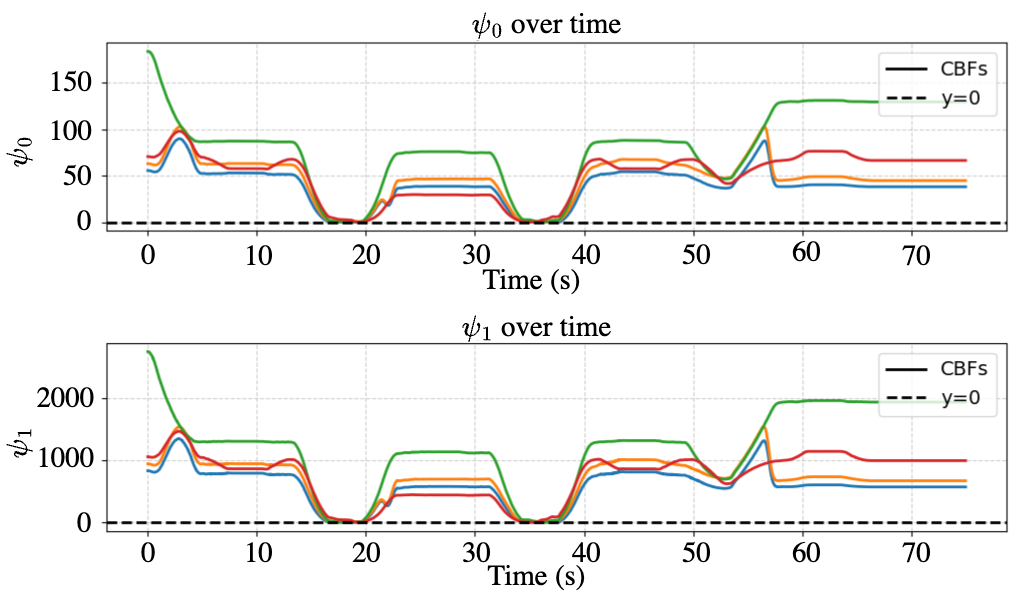}
    \caption{Time histories of four selected HOCBFs for the inter-robot safety constraints in the pick-and-place example. The top panel shows $\psi_0$s of the HOCBFs and the bottom panel shows $\psi_1$s of the HOCBFs. All barrier conditions remain satisfied throughout the trajectory.}
    \label{fig:cbf_hist}
\end{figure}

The proposed algorithms were conducted on two Franka Research 3 arms mounted in proximity of each other, as shown in Fig.~\ref{fig:real_world_experiment environment}. Similar to simulation experiments, four ellipsoids per robot were used to define HOCBFs, resulting in a total of 16 HOCBFs. The proposed centralized, decentralized, and relaxed-decentralized safety filters, incorporating both analytically computed and estimated \emph{Hessian contributions}, were tested on 10 different intersecting trajectories. Additionally, a collaborative pick-and-place task was designed to illustrate the robots' ability to collaborate safely under the proposed safety methodology. One such experiment using the centralized filter with estimated \emph{Hessian contribution} is shown in Fig.~\ref{fig:real_robot}. The deviation between the nominal control input and the safe filtered input is shown in Fig.~\ref{fig:nominalvsfiltered}, and the time histories of four representative HOCBFs are provided in Fig.~\ref{fig:cbf_hist}. These figures show that the filtered inputs diverge from the nominal inputs primarily when the HOCBF values approach zero, indicating that the safety filter activates only when necessary.

In these experiments, the window length and polynomial order of the Savitzky-Golay filter used to estimate the \emph{Hessian contribution} were set to 5 and 2, respectively. The experiments were conducted on a computer equipped with a 24-core Intel Core i9-13900K processor and 64 GB of RAM. All filters, whether based on computed or estimated \emph{Hessian contributions}, successfully executed all experiments while maintaining safety. On average, the filters using analytically computed \emph{Hessian contributions} operated at up to 150 Hz. Although the Savitzky-Golay filter assumes a constant sampling rate and system interruptions caused variations in the control period beyond the 275 Hz control update rate, the safety filters using estimated \emph{Hessian contributions} could achieve an average operation rate of 400 Hz without compromising robot safety.

When we conducted the experiments on a less powerful PC with a 16-core Intel Core i7-11700 processor and 64\,GB of RAM, analytically computing the \emph{Hessian contributions} resulted in a control update rate of 50 Hz, which is too slow to run the robots in our experiments. However, the estimated \emph{Hessian contributions} still operated at 135 Hz on average, which was sufficient to run the robots in our experiments. Given that the estimated \emph{Hessian contributions} maintained safety in all the experiments, the proposed estimation method is a valid alternative when high control frequencies are required or when computational constraints prevent analytical computation.
\section{Conclusion}
We presented HOCBF-based safety constraints to prevent inter-robotic arm collisions. Using these constraints, we developed centralized, decentralized, and relaxed-decentralized filters. We addressed the computational overhead of HOCBFs by introducing an efficient estimation method based on numerical derivation. We validated the effectiveness of the proposed safety approach through simulation and real-robot experiments. In these experiments, we observed that decentralized filtering can suffer from QP infeasibility in certain scenarios. Although relaxation mitigated this issue, it did not fully resolve it. Future work will focus on addressing this issue and applying the method in more challenging environments, including mobile robots with robotic arms.
\section*{CRediT authorship contribution statement}
\textbf{Ali Umut Kaypak}: Methodology, Writing-Original Draft, Software, Conceptualization, Writing - Review \& Editing, Visualization. \textbf{Shiqing Wei}: Methodology, Software, Conceptualization, Writing - Review \& Editing. \textbf{Prashanth Krishnamurthy}: Methodology, Conceptualization, Writing - Review \& Editing, Supervision, Funding acquisition. \textbf{Farshad Khorrami}: Methodology, Conceptualization, Writing - Review \& Editing, Supervision, Funding acquisition. 
\section*{Declaration of competing interest}
The authors declare that they have no known competing financial interests or personal relationships that could have appeared to influence the work reported in this paper.
\section*{Data availability}
Data will be made available on request.
\section*{Acknowledgment}
This work was supported in part by the Army Research Office under grant numbers W911NF-21-1-0155 and W911NF-22-1-0028 and New York University Abu Dhabi (NYUAD) Center for Artificial Intelligence and Robotics (CAIR), funded by Tamkeen under the NYUAD Research Institute Award CG010.
\bibliographystyle{elsarticle-num}
\bibliography{references}

\end{document}